# Path Planning of an Autonomous Mobile Robot in a Dynamic Environment using Modified Bat Swarm Optimization

Ibraheem Kasim Ibraheem[1], Fatin Hassan Ajeil[1], Zeashan H. Khan[2]

[1]Electrical Engineering Department, College of Engineering, University of Baghdad, AL-Jadriyah, P.O.B.: 47273, 10001 Baghdad, Iraq.

[2]Department of Electrical Engineering, Bahria University, Islamabad, Pakistan, 44000

**Abstract:**

This paper describes the path planning of an autonomous mobile robot using an advanced version of swarm optimization using Bat Algorithm (BA) in a dynamic environment. The main objective of this work is to obtain a collision-free, shortest, and safest path between starting point and end point assuming a dynamic environment with moving obstacles. A new modification on the frequency parameter of the standard BA has been proposed in this work, namely, the Modified Frequency Bat Algorithm (MFBA). The path planning problem for the mobile robot in a dynamic environment is carried out using the proposed MFBA. The path planning is achieved in two modes: (1) the first mode is the path generation and is implemented using the MFBA, this mode is enabled when no obstacles near the mobile robot. When an obstacle close to the mobile robot is detected, (2) the second mode, i.e., the obstacle avoidance (OA) is initiated. Simulation experiments have been conducted to check the validity and the efficiency of the suggested MFBA based path planning algorithm by comparing its performance with that of the standard BA. The simulation results showed that the MFBA outperforms the standard BA by planning a collision-free path with shorter, safer, and smoother than the path obtained by its BA counterpart.

## 1. Introduction:

Mobile robot navigation is a challenging problem in the robotics field and numerous studies have been endeavored resulting in a considerable number of solutions [1]. The term navigation refers to the guidance of the mobile robot from the starting position to the target position avoiding collisions and unsafe conditions [2]. The research of path planning began in the late 60's of the past century, and several algorithms have been proposed. These comprise the roadmap method [3], cell decomposition [4], Potential fields [5], and mathematical programming [6], etc., just mentioning a few. Lately, it is discovered that these algorithms are either ineffective, because of the significant computational cost; or imprecise, because of getting stuck in local minimum. To outdo these drawbacks, several heuristic based methods have been implemented, such as the application of artificial neural networks, particle swarm optimization (PSO), genetic algorithm (GA) [7], and hybridization between them [8]. One of the main benefits of heuristic based methods is that it can yield satisfactory results quickly, which is particularly appropriate to solve NP-complete problems.

The path planning is separated into two main fields, global and local path planning. On the first hand, in path planning with local path planning, the calculations of the path are achieved when the mobile robot is in motion; that means, the calculation is fit for generating new paths as the environment changes. On the other hand, with global path planning, the environment should be totally recognized and identified, while the terrain must be static [9,10]. Many types of research have been studied path planning problem in dynamic environments, authors of [11] proposed a new method to decide the optimum route of the mobile robot in an unknown dynamic environment by the ant colony optimization (ACO) algorithm, they used ACO algorithm to decide the optimal rule table of the fuzzy system. The angle variance to the nearest obstacle and the distance between the starting and end positions are the factors that affected the fuzzy decision-making process. The work in [12] presented the navigation approach for ground vehicle in a dynamic environment by using two types of fuzzy logic controllers. The first controller called the Target Reaching fuzzy Inference Controller (TR-FIC) to guarantee to have the robot reaches the goal (by means of the angle difference between vehicle heads and target). The second controller is called an Obstacle Avoidance Fuzzy Inference Control (OA-FIC). The switching between these controllers is done by obstacle sensing bit from the environment. While in [13] researchers introduced a new method to deal with the static and dynamic obstacles, if the robot encounters a static obstacle, it avoids the obstacle by using a fuzzy logic controller. While in the case of dynamic obstacles, the mobile robot estimates the direction and the velocity of all the moving obstacles and generates a corresponding trajectory prediction table of each obstacle for estimating the obstacle's future trajectory. In [14], people proposed a new fitness function taking into account the distances between robot-obstacles, robot-robot, and robot-goal. This work deals with obtaining the optimal path for multiple mobile robots using PSO algorithm. In [15] obstacle avoidance for wheeled mobile robot in a dynamic environment is achieved by considering a virtual disc in front of the mobile robot centered at the head angle of the mobile robot, then computing the intersection angle between the disc and the obstacles and changing the direction of the mobile robot according to specific rules. The fuzzy logic system and laser scan sensor were implemented in [16] for tuning mobile robot velocities in an unknown dynamic environment, the noise of the laser sensor has been eliminated with a suitable filter. The movement of the dynamic obstacles has been predicted using artificial neural networks (ANN) and Radial Basis Functions Neural Networks (RBANN) to solve the problem of the motion planning in [17]. While in [18] a new method has been suggested for tackling the motion planning problem in a dynamic environment using the synergism of neural network, and fuzzy system. This work combined Artificial Potential Field (APF) and Fuzzy Neural Network (FNN) for globally optimized path planning on a dynamic environment in a real-time fashion. This technique used Fuzzy Logic to construct the space model, then integrated APF and FNN exploring for the shortest route while efficiently evades moving obstacles. Grid-based methods for mobile robot path planning can be found in [19]. The global search speed of bat algorithm has been enhanced based on mutation between bats during updating bats position, this enhancement capability applied to Uninhabited Compact Air Vehicle (UCAV) in [20], while in [21] hybridization between bat algorithm and cuckoo search proposed to reduce iterations number and simulation time to find shortest path. It should be remembered that mobile robot navigation including the path planning can be considered as the upper layer for the motion planning of the mobile robot through which the mobile receive data and react to its environment. This layer is built on the lower layer, namely, the motion control layer which operates the actuators of the mobile robot in response to the upper layer. Motion control layer can be designed using one of the linear or nonlinear control design methods [22–28].

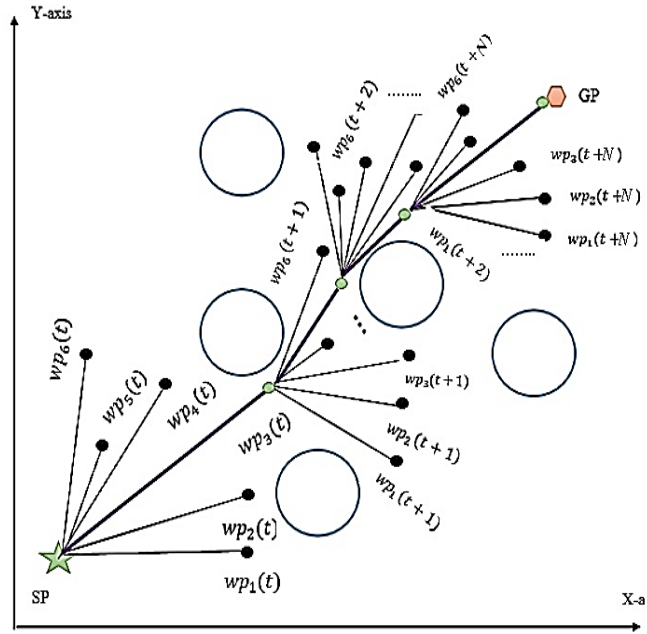

**Fig. 1. Dynamic Environment representation for mobile robot path planning.** $wp_n(t)$ is the nth point of iteration $t$, $N$ is the number of iterations, the black empty circles are dynamic obstacles. The green points are the best points (that give the shortest distance to GP) selected by the proposed algorithm at each iteration.

The main contribution of this paper is developing a new path planning algorithm which consists of two main modules: 1) the first module involves point generation, achieved using a novel heuristic nature-inspired algorithm, namely, the Modified Frequency Bat Algorithm (MFBA). The MFBA generates and select the points that satisfy the measure adopted in this work, i.e., the shortest path between the start and goal points. It finds a collision-free and shortest path (if exists) from an initial position to an end position in a dynamic environment with variable speeds moving obstacles, 2) next the MFBA is integrated with a second module in which a local search technique detects dynamic obstacles. Moreover, to avoid these obstacles, twelve sensors are deployed around the mobile robot to sense their existence, and once are detected, an avoidance algorithm is triggered and a new collision-free position for the mobile robot is found to resume its trip to its goal position based on gap vector principle.

The rest of this paper is organized as follows. Section 2 presents the problem statement and assumptions. The swarm based navigation algorithm is introduced in Section 3. While Section 4 suggests the obstacle detection and avoidance algorithm. The simulations and the results are discussed in Section 5. Finally, the conclusions are given in Section 6.

## 2. Problem Statement and Assumptions

Assuming a 2-D world frame as shown in Fig.1 above, the mobile robot is at initial point (SP) and has to reach the final point (GP). There are dynamic obstacles in the environment. The objective is to find the shortest route from SP to GP without colliding with any of the dynamic obstacles in the workspace by finding the best and feasible next position for the mobile robot from the current one. Before discussing and suggesting the solution to this problem, there are some assumptions made in this work:

- The obstacles and mobile robot are represented by an equally size circular shape.
- No kinematic constraints affect the motion of the mobile robot. The only effect source is the motion of the obstacles.
- The speed of the obstacles is fixed and different.
- The mobile robot movement is omnidirectional at any time.
- The shortest distance is obtained by minimizing the distance function $f(x,y)$,

$$f(x,y) = \sqrt{(x_{i+1} - x_i)^2 + (y_{i+1} - y_i)^2} \quad (1)$$

$$fitness = \frac{1}{f(x,y)+\varepsilon} \quad (2)$$

where: $x_{i+1}, y_{i+1}$ represent the next position, $x_i, y_i$ represent the current position, and $\varepsilon$ is small number (say 0.001) to prevent division by zero.

## 3. SWARM BASED NAVIGATION ALGORITHM

The BA is a bio-inspired algorithm developed by Yang in 2010, it is based on the echolocation or bio-sonar characteristics of micro-bats [29-30]. Echolocation is an important feature of the bat behavior, which means the bats emit a sound pulse and listen to the echo bouncing back from the obstacles while flying.

### 3.1. Bat Algorithm (BA)

By utilizing the time difference between its ears, the loudness of the response, and the delay time, the bats can figure up the velocity, the shape, and the size of the prey. In addition, the bat has the capability to change the way it works, if it sends the sound pulses with a high rate, they won't fly longer but give thorough details about its nearby surroundings which help bats to distinguish the prey position exactly. Another characteristic of bat's echolocation is its loudness; when the bats are near from the victim, it transmits sound pulsations silently while amid the searching process they send noisy sound pulses. Bats hunting methodology can be summarized in the next:

All the bats utilize the echolocation to detect the distance. In addition, they distinguish the distinction between nutrient/prey and background obstacles in some supernatural way. Bats fly at random with speed vi at posture xi with a frequency $f_{min}$, changing wavelength λ and loudness $A_0$ to scan for the prey. They can consequently change the wavelength (or frequency) of their transmitted pulsations and alter the rate of pulses transmission r ∈ [0, 1], contingent upon the nearness of their objective. Even though the loudness can change in many ways, we presume that the loudness decreases from a large (positive) $A_0$ to a minimum constant value $A_{min}$.

#### 3.1.1. The Movement of Artificial Bats

In a D-dimensional searching or solution space is basically a region in which prey/food is found. So bat move randomly around the search space to find the prey since they have no idea about the prey. The quality of the food is determined by the fitness of bat. Initial population for n number of bats is generated randomly by the following equation:

$$x_{i,j} = x_{min,j} + rand * (x_{min,j} - x_{max,j}) \qquad (3)$$

where each bat is related with a velocity $v_i(t)$ and a location $x_i(t)$ at iteration '$t$'. Amid all the bats, there exists a current best solution. As a result, the guidelines aforementioned above can be converted into the updating equations for the positions and velocities. Since, the frequency $f_i$ controls the range and the space of the movement, the updating procedure of bat's positions/solutions is as follows:

$$f_i = f_{min} + (f_{max} - f_{min}) * β \qquad (4)$$

$$v_i(t) = v_i(t-1) + (x_i(t-1) - x*)f_i \qquad (5)$$

$$x_i(t) = x_i(t-1) + v_i(t) \qquad (6)$$

where β ∈[0, 1] is a random vector of a uniform distribution. Here $x*$ is the present global best location (solution), which is found after comparing all the solutions among all the n bats. For the locally searching stage, once a solution is chosen from the best current solutions, a new solution for each bat is locally produced using the random walk principle:

$$x_{new} = x_{old} + σϵA(t) \qquad (7)$$

where $\epsilon \in [-1, 1]$ is a random number and represents the direction and intensity of random walk, A (t) is the average loudness of all the bats at iteration step t. From the practical point of view, it is better to provide a scaling parameter σ to control the step size.

### *3.1.2. Loudness and Pulse Emission*

The loudness $A_i$ and the rate of pulse emission $r_i$ have to be updated accordingly as the iterations continue. Typically, the loudness $A_i$ decreases once a bat has detected its prey, whereas the rate of pulse emission $r_i$ rises according to the following equations:

$$A_i(t + 1) = \alpha A_i(t) \tag{8}$$

$$r_i(t + 1) = r_i(0)[1 - e^{-\gamma t}] \tag{9}$$

where α and γ are constants. For any 0 < α < 1 and γ > 0. Notwithstanding the success of the BA and its diverse fields and applications, there are still a few fundamental issues that require more investigation as described next.

## 3.2. Modified Frequency Bat Algorithm (MFBA)

There are two significant divisions in current metaheuristics: Exploration and exploitation. Exploration is the inspection ability of mysterious different spaces to sense the global optimal target, while exploitation points to finding the optimal target by utilizing the past best fitness's information. A decent reply of a metaheuristics algorithm relies upon good coordination of these modules. In the case of little Exploration with escalated exploitation, the procedure could be stuck into local optima [31]. While a considerable Exploration with little exploitation could bring the algorithm to converge gradually and reduces the overall searching performance. The BA is an effective optimization algorithm in "exploitation" (i.e., local search), but at certain times it may get stuck at local optima and consequently it cannot accomplish the global search efficiently. For BA, the searching depends totally on "random walks", so a rapid convergence cannot be secured [32].

Since the frequency controls the space and the range of the Bat movement, the sound pulses with high frequency will not travel longer and vise verse. In addition, the BA loses its exploration capability when through time. Therefore, we suggest a method for balance between exploration and exploitation using frequency tuning by assuming that the algorithm starts with low frequency to increase the global search capability and the frequency increases gradually as iterations proceed. This can be achieved by considering the factor β as a function; instead of being a random number. In this way, it guarantees that the frequency will increase through iterations. The factor β in (3) will be modified as follows:

$$\beta = t * e^{(-\rho * r)} \tag{10}$$

where $t$ is iteration number, r is a random number (0,1), the value of ρ is an application dependent, for path planning problem the value of ρ is chosen to be (0.01) as shown in the Fig. 2.

## 4. OBSTACLE DETECTION AND AVOIDANCE: LOCAL SEARCH

The mobile robot navigates from its SP to GP using BA or MFBA until it detects an obstacle, then the mobile robot switches to local search mode. The local search is implemented by surrounding the mobile robot by twelve virtual sensors as shown in Fig. 3.

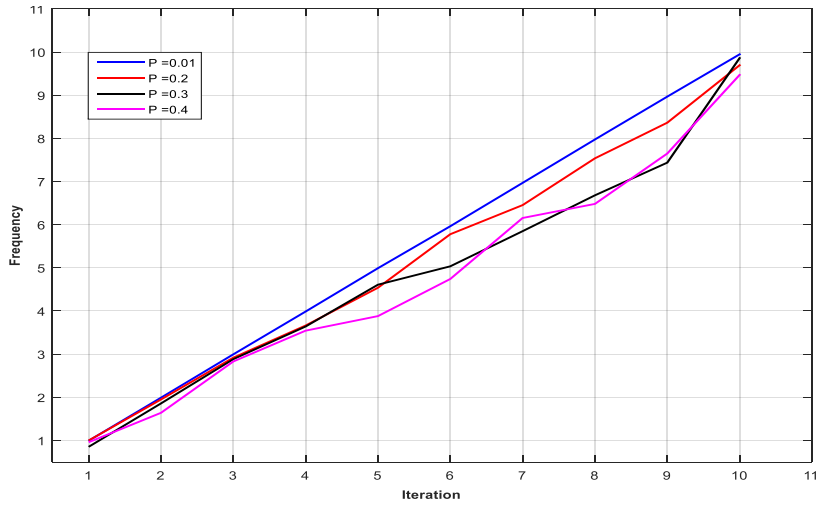

**Fig. 2. Increasing frequency parameter of BA with iterations and different values of ρ**

Since the mobile robot is a physical body, the obstacles are enlarged by the dimension of the mobile robot and considering the robot as a particle during the simulations as depicted in Fig.4.

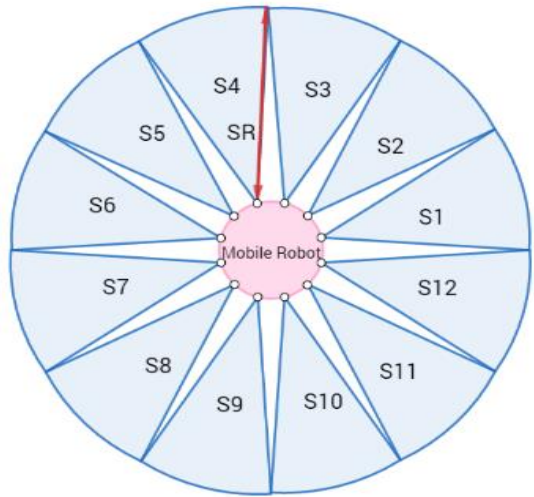

**Fig. 3. Mobile robot sensors deployment**

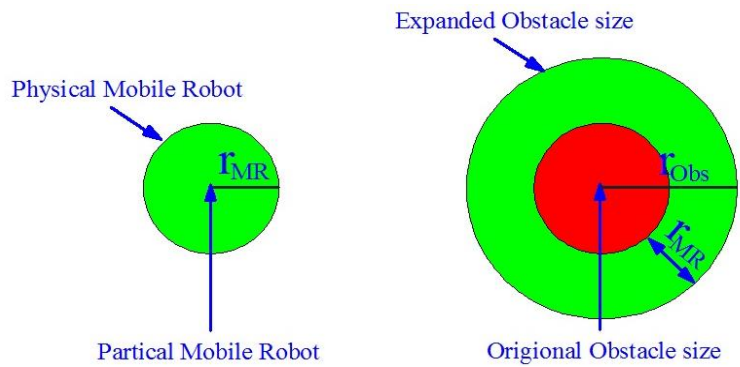

**Fig. 4. Enlarging obstacles size corresponding to mobile robot size, $r_{MR}$ is the mobile robot radius, while $r_{Obs}$ is the radius of the expanded dynamic obstacle.**

*4.1. Obstacles Detection*

The obstacles are detected by using *sensory vector* (binary vector provides information about the existence of obstacles, its dimension depends on number of sensors (in this case 12 sensor used), therefore, the sensory vector is represented by twelve bit indexed by $i \in \{1,2,...,12\}$.

$$VS = [a(1), a(2), ..., a(i), ..., a(12)] \tag{11}$$

The value of each bit indicate the existence of obstacle ($a(i) = 1$) or not ($a(i) = 0$). These sensors have equally Sensing Range $SR$ (the maximum distance the mobile robot can measure) and angle of sensor ($\frac{360°}{no\ of\ sensor}$). For each obstacle inside a certain sensor range (i.e. in the range between $Si$) draw the tangent lines to the expanded obstacle (dotted circles in Fig. 5). Then compute the angle between the mobile robot point and the obstacle tangent lines that belong to a certain sensor range $Si$. For example, consider Fig. 5, for obstacle ($Obs_1$) the range of tangent angles from $T_{P1}$ to $T_{P2}$ belong to $S1$ and $S2$ so, the corresponding bits ($a(1)$, $a(2)$) in $V_S$ set to logic (1), same procedure applied to Obstacles ($Obs_3$, and $Obs_4$) in Fig.5, the sensory vector Vs is given as:

$Vs = \{1\ 1\ 0\ 0\ 0\ 0\ 1\ 1\ 1\ 0\ 0\ 0\}$

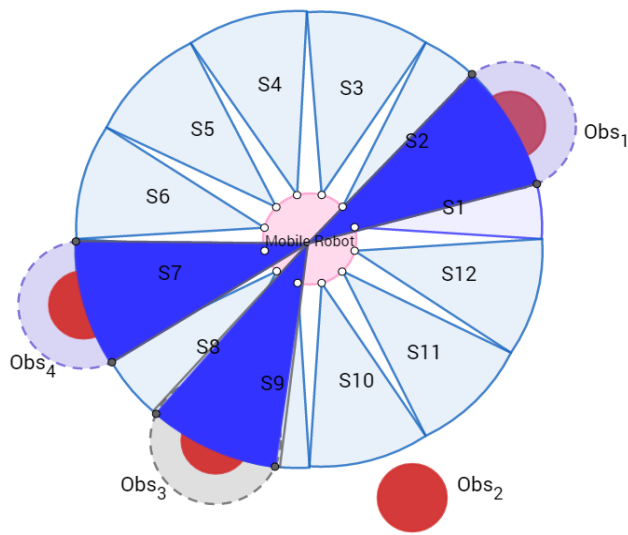

**Fig. 5. The detection, sensing and angle ranges of the mobile robot.**

*4.2. Obstacles avoidance*

The obstacles avoidance is done by using gap vector concept. Gap vector Vg, which is a binary vector, where logic(1) represent the occupancy gap, while logic (0) represent a free gap. The length of the vector is given as:

$$\text{length } (Vg) = \text{length } (Vs) \tag{12}$$

$$Vg = \{b(1), ..., b(i), ..., b(12)\} \tag{13}$$

The mobile robot chooses the nearest gap to the goal. The gap vector Vg can be derived from the sensing vector Vs (each consecutive zeros represent free gap (logic( 0)), otherwise ( logic (1) ) which is equivalent to OR gate as shown in Table 1 and Fig.6.

**Table 1. Gap Vector Construction**

| Vs(i) | Vs(i+1) | Vg(i) |
|---|---|---|
| 0 | 0 | 0 |
| 0 | 1 | 1 |
| 1 | 0 | 1 |
| 1 | 1 | 1 |

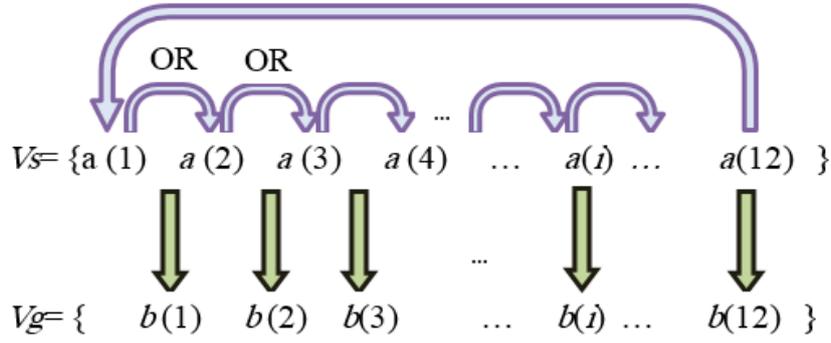

**Fig. 6. Graphical illustration of Gap Vector Construction**

From the previous example (Fig. 5) we get,

$Vs = \{1\ 1\ 0\ 0\ 0\ 0\ 1\ 1\ 1\ 0\ 0\ 0\}$

$Vg = \{1\ 1\ 0\ 0\ 0\ 1\ 1\ 1\ 1\ 0\ 0\ 1\}$

## 5. Proposed Path Planning Algorithm

The proposed path planning algorithm can be clarified in the Pseudo-Code listed in Algorithm 1.

---
**Algorithm 1**: Pseudo code for Proposed modified BA
---

*while* (mobile robot position is not equal to GP)
   *collect data from virtual twelve sensors*
   *if detected obstacles within SR*

   *switch to local search mode (OA mode)*

*else*
   *navigate toward goal using MFBA of* (3)-(9)
  *end if*

*end while*

---

## 6. Simulation Results

The simulations of the proposed MBFA is divided into two parts. The first part includes the simulation of the test of the MFBA on benchmark functions. While the second one is devoted to the simulations for the path planning of the mobile robot on a dynamic environment.

### 6.1. Standard Benchmark Functions

The assessment of validation, efficiency, and reliability of a certain optimization algorithm is normally implemented by utilizing a set of basic standard benchmarks or test functions. Then the performance of such an algorithm is approved by comparing it with its counterparts after utilizing these test benchmark functions in the calculations. Among the most popular ones are given in Table 2. The proposed MFBA has been tested with 15 runs for each benchmark function. In order to analyze the performance of both algorithms, the "mean" values in Table 3 are considered. The sign "+" indicates that MFBA is better than BA, "-" means that the two approaches relatively give equal results, and "." represents MFBA is worse than BA.

**Table 2. Standard Benchmark Functions**

(C: characteristic, U: unimodal, M: multimodal, d: dimension, m: steepness parameter =10)

| Fun Name | Formula | C | Fmin | Search range |
|---|---|---|---|---|
| F1: Sphere | $\sum_{i=1}^{d} x_i^2$ | U | 0 | [-5.12, 5.12] |
| F2: Easom | $-\cos(x_1)\cos(x_2) * \exp(-(x_1-\pi)^2 - (x_2-\pi)^2)$ | U | -1 | [-100, 100] |
| F3: Three Hump Camel | $2x_1^2 - 1.05x_1^4 + \frac{x_1^6}{6} + x_1 x_2 + x_2^2$ | M | 0 | [-5,5] |
| F4: Booth | $(x_1 + 2x_2 - 7)^2 + (2x_1 + x_2 - 5)^2$ | U | 0 | [-10,10] |
| F5: Rastrigin | $10 * d + \sum_{i=1}^{d}(x_i^2 - 10 \cdot \cos(2 \cdot \pi \cdot x_i))$ | M | 0 | [-5.12, 5.12] |
| F6: Michalewicz | $-\sum_{i=1}^{d} \sin(x_i) \cdot \sin^{2m}(\frac{i \cdot x_i^2}{\pi})$ | M | -1.8013 | [0, π] |

**Table 3. The comparison performance of the BA and MFBA on benchmark functions**

| Fun No | Alg. | Best | Worst | Mean | SD | significant |
|---|---|---|---|---|---|---|
| F1 | BA | 0.00028437 | 0.048626 | 0.0146 | 0.0145 | |
| | MFBA | 0.00011976 | 0.0285 | 0.0048 | 0.0078 | + |
| F2 | BA | -0.8286 | -9.64e-32 | -0.1241 | 0.2476 | |
| | MFBA | -0.8691 | 0 | -0.2090 | 0.3132 | + |
| F3 | BA | 1.1407e-05 | 0.3050 | 0.0398 | 0.0857 | |
| | MFBA | 1.1183e-05 | 0.1034 | 0.0154 | 0.0361 | + |
| F4 | BA | 1.7676e-04 | 0.4544 | 0.1256 | 0.1533 | |
| | MFBA | 2.7530e-05 | 0.1854 | 0.1029 | 0.0305 | + |
| F5 | BA | 0.0121 | 5.6793 | 2.4541 | 1.8399 | |
| | MFBA | 0.0642 | 5.3864 | 2.0278 | 1.3476 | + |
| F6 | BA | -1.9880 | -1.5022 | -1.7743 | 0.1008 | |
| | MFBA | -1.9695 | -1.7993 | -1.8762 | 0.0617 | + |

*6.2. Simulation Results on Mobile Robot Path Planning*

The simulation parameters for all case studies are population size = 5, $A_i(0)= 1$, $r_i(0)= 0.5$, $\alpha = 0.98$, $\gamma = 0.8$, $f_{min} = 0$, $f_{max}=10$, $\sigma = 0.3$, SR = 0.8, the starting point for mobile robot is SP = (0, 0), and goal is GP = (12, 12). The simulations have been run under MATLAB Environment on a computer system with 2.76 GHz Core i7 CPU, and 4G RAM. In these case studies, the solutions (shortest paths) are obtained after executing the proposed algorithm ten times in order to find out the best path. In all the simulations, the optimal path from start to target points without any obstacles is equal to 16.9705. The best path is the one nearest to the optimal path.

- **Case Study 1: Three dynamic obstacles.**

In this case study, the positions, velocities, and directions of the obstacles are listed in Table 4 below:

**Table 4. Characterization of the obstacles motion**

| Obs | center | radius | Vobs(m/s) | Theta(deg) |
|-----|--------|--------|-----------|------------|
| 1 | (1, 4.5) | 0.3 | 0.3 | 0° |
| 2 | (10.5, 6) | 0.3 | 0.2 | 180° |
| 3 | (6, 12) | 0.3 | 0.15 | 270° |

The mobile robot navigates from (0, 0) as a start point using MFBA, the magenta circle around the mobile robot in Fig. 7 (a) represents the sensing region. When the moving obstacles being sensed in this region, the path planning algorithm is switched into the local search mode and consequently triggers OA procedure as shown in Fig.7 (b). Then the mobile robot continues its searching process until it reaches the goal as shown in Fig.7 (c, d). The best path with the shortest distance and shortest time using MFBA algorithm was achieved in experiment no. eight among the ten runs. The total distance was equal to 17.0925m with a run time equal to 8.151829 sec. While the best path with the shortest distance using BA algorithm was achieved in experiment no. 8 too. The total distance was equal to 17.0932m with a run time equals 7.905828 sec.

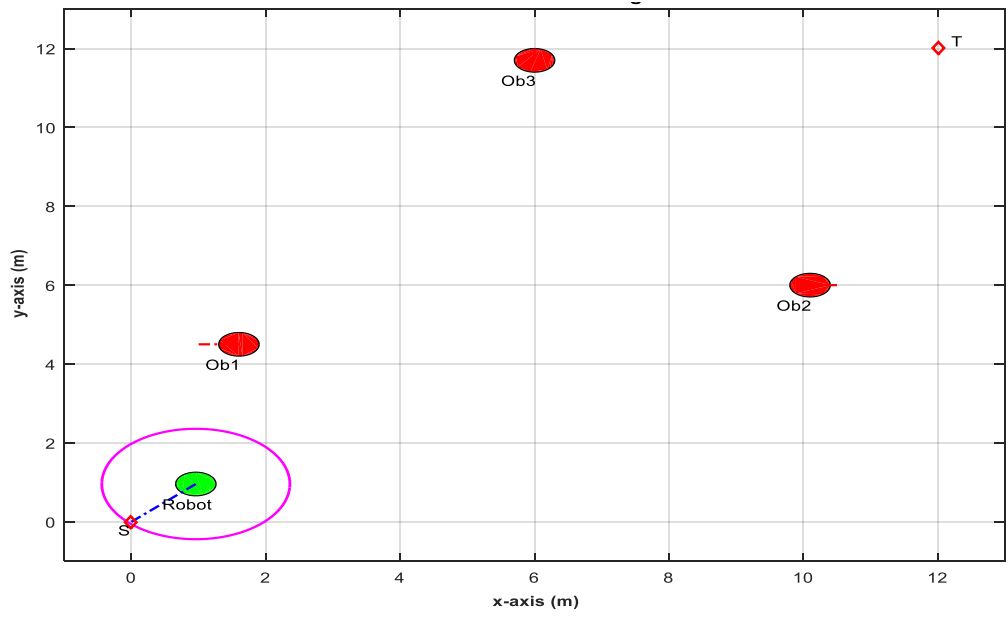

(a): No obstacles around mobile robot

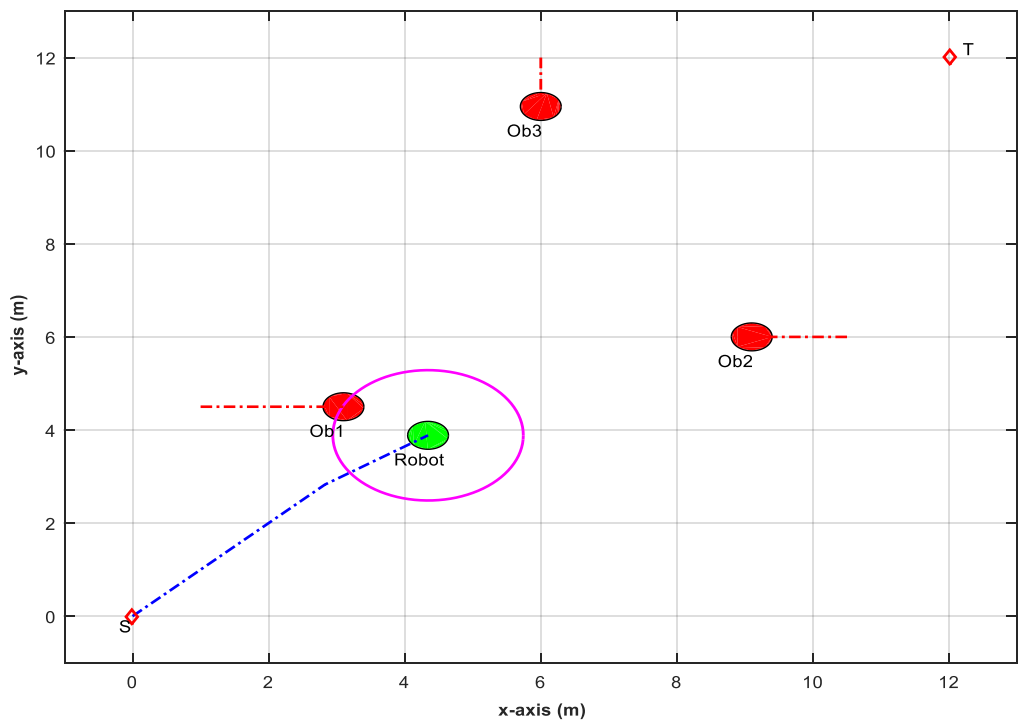

(b) : Ob1 enter sensing region

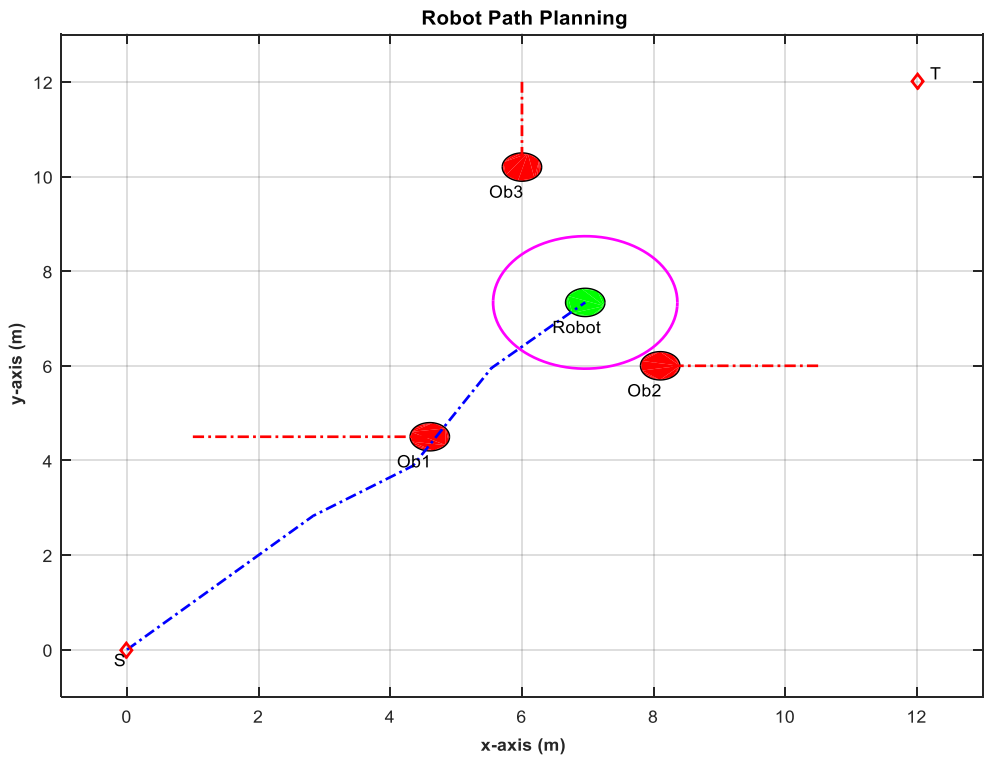

**(c): mobile robot returns to normal mode**

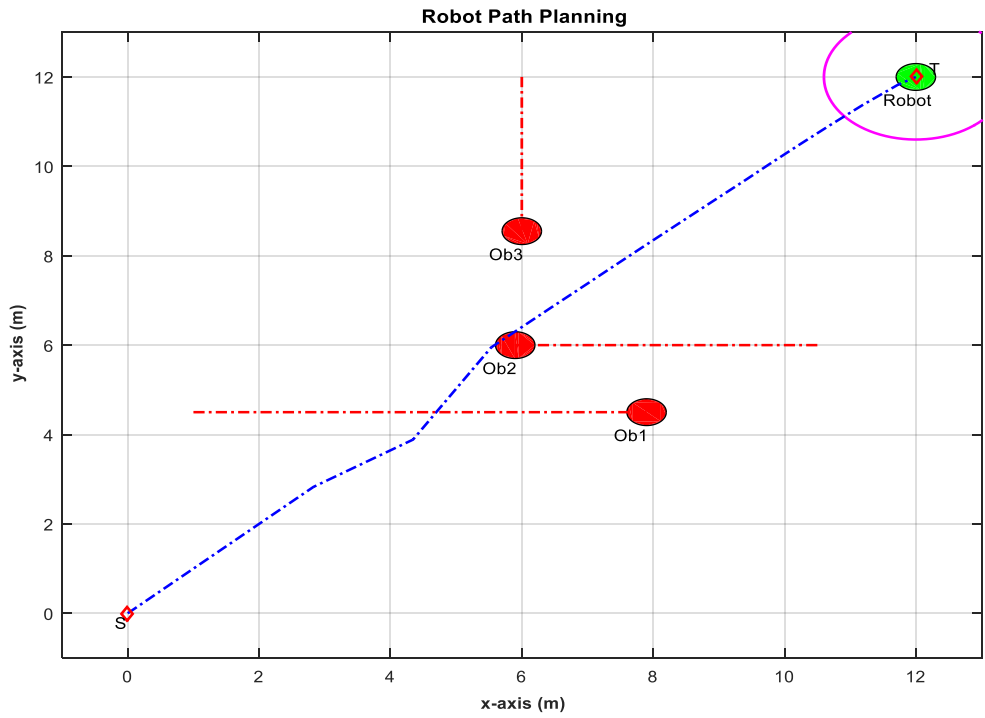

**(d)**: mobile robot arrives at its goal

**Fig. 7. The best path for case study 1 with three dynamic obstacles.**

Another comparison is made by summarizing both algorithms results after executing the program ten times. The MFBA obtained the largest fitness as the smallest standard deviation achieved by BA, the results are tabulated in Table 5.

**Table 5. Comparison results for case study 1**

| Fitness | Standard BA | MFBA |
|---|---|---|
| minimum | 0.056170939 | 0.05830325 |
| maximum | 0.058502796 | 0.0585072461 |
| Standard deviation | 0.00103782 | 0.05315157 |
| mean | 0.05751639825 | 0.05844756632 |

- **Case Study 2: Five dynamic obstacles**

In this case study, the positions, velocities, and directions of obstacles are listed in Table 6 below.

**Table 6. Characterization of the obstacles motion**

| Obs | center | radius | $V_{obs(m/s)}$ | Theta(deg) |
|---|---|---|---|---|
| 1 | (4, 2) | 0.3 | 0.3 | 111.8° |
| 2 | (3,7) | 0.3 | 0.2 | 315° |
| 3 | (9, 4) | 0.3 | 0.2 | 126.8° |
| 4 | (7, 9) | 0.3 | 0.25 | 315° |
| 5 | (11.2,7) | 0.3 | 0.22 | 150° |

The mobile robot navigates in between five dynamic obstacles with different velocities as shown in Fig.8, where the red dotted lines refer to the direction of the obstacle's movement. It's initially moving toward GP using MFBA until two of the obstacles ( the first is shown in Fig. 8 (b), while the second is shown in Fig. 8 (c) enter the sensing region. At that time, the local search mode has been enabled to avoid obstacles as shown in Fig. 8 (b, c), the mobile robot returns to its normal mode (i.e., path generation using MFBA or BA) as shown in Fig. 8 (d, e). The best path with the shortest distance and shortest time using MFBA algorithm was achieved in experiment no. five among the ten runs. The total distance was equal to 18.3533 m with run time 9.0903 sec. While the best path using BA algorithm was achieved in experiment no. 8, which was equal to 18.6239 m with a run time equal to 9.10554 sec.

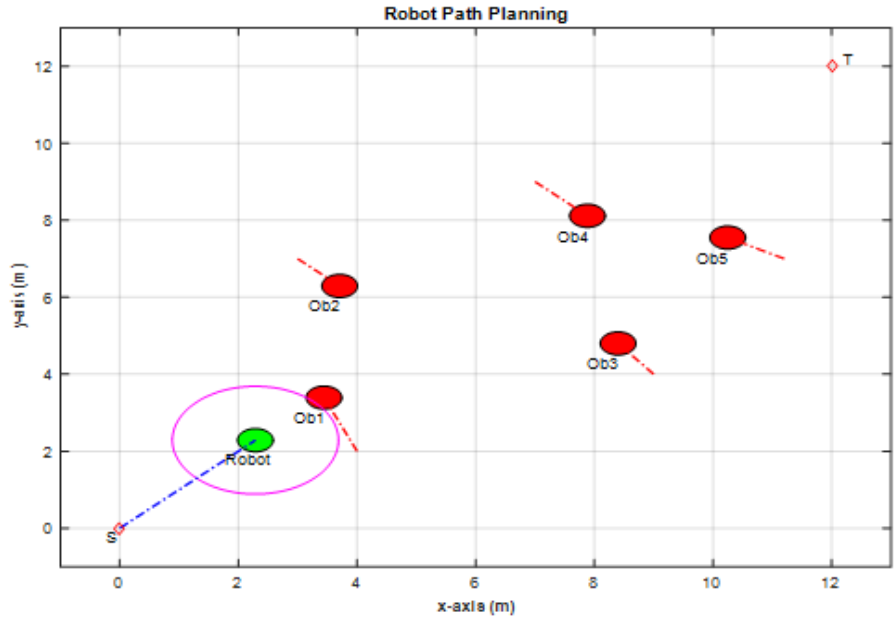

(a): no obstacle around mobile robot

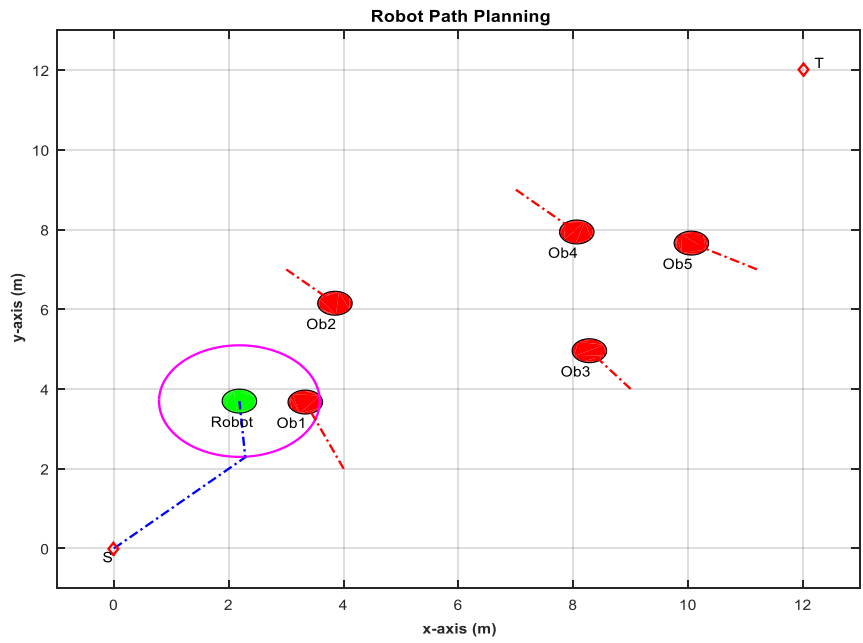

(b): obstacle (ob1) enter sensing region

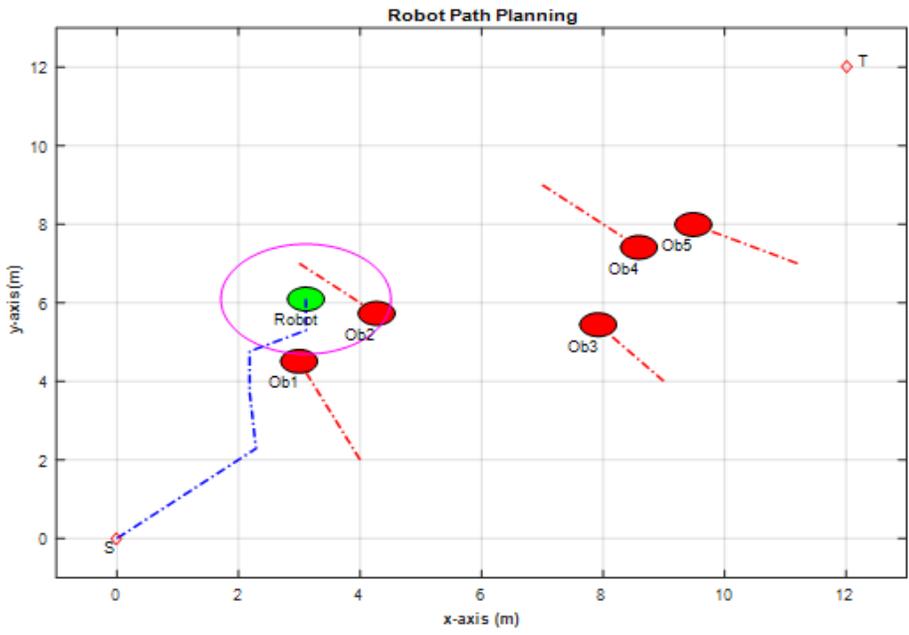

(**c**): obstacles (ob1, ob2) inside the sensing region

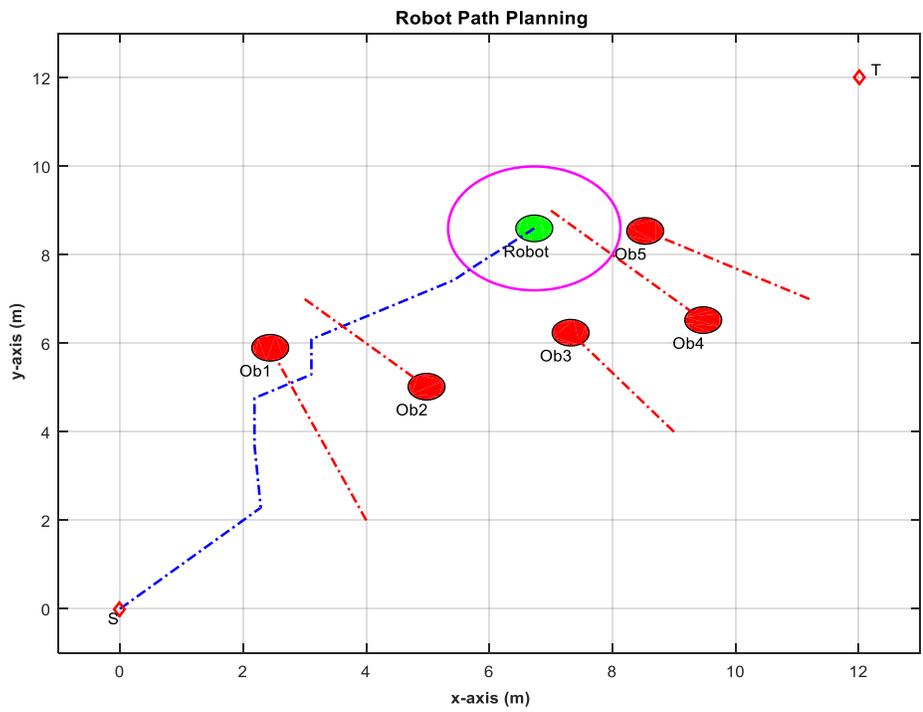

**d**): obstacles (ob1, ob2) return to normal

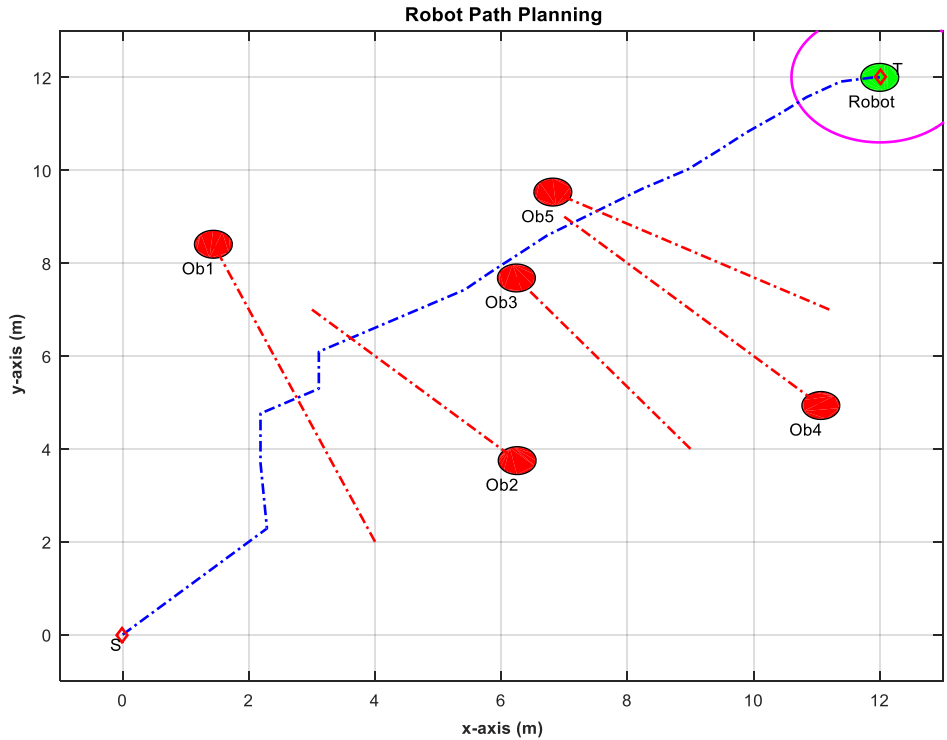

(**e**): mobile robot arrives at its goal
Fig. 8. The best path for case study 1 with three dynamic obstacles.

Finally, a comparison is made by summarizing the results of both algorithms after executing the program ten times. The MFBA obtained the largest fitness with the smallest standard deviation as tabulated in Table 7.

Table 7. Comparison results for case study 2

| Fitness | Standard BA | MFBA |
|---|---|---|
| minimum | 0.052038612 | 0.0532810468 |
| maximum | 0.053694446 | **0.054486114** |
| Standard deviation | 0.10387507 | **0.07777658** |
| mean | 0.052918954 | 0.05399353627 |

## 7. Conclusion

In this paper, the navigation approach for the mobile robot in a dynamic environment has been introduced by using MFBA optimization combined with the local search technique and obstacle avoidance. The modified algorithm allowed the mobile robot to move from its start location to destination without colliding any of the available moving obstacles in the dynamic environment. The modification of the frequency parameter in the standard BA guaranteed that the searching was more directed and faster. The simulations on the test benchmark functions together with that on the path planning of a mobile robot proved the validity and the effectiveness of the proposed MFBA as compared to the standard BA. It can be concluded that the proposed MFBA based path planning achieves the optimal goal with the shortest and path as compared to its BA based path planning counterpart. We suggest applying these techniques over a real test bed of mobile robots. Moreover, the algorithm should also be tested for the kidnapped robot problem.